\documentclass[sn-mathphys,Numbered]{sn-jnl}
\usepackage{hyperref}
\usepackage{graphicx}
\usepackage{multirow}
\usepackage{amsmath,amssymb,amsfonts}
\usepackage{amsthm}
\usepackage{mathrsfs}
\usepackage[title]{appendix}
\usepackage{xcolor}
\usepackage{textcomp}
\usepackage{manyfoot}
\usepackage{booktabs}
\usepackage{algorithm}%
\usepackage{algorithmicx}
\usepackage{algpseudocode}
\usepackage{listings}

\begin{document}

\title[Article Title]{Research on Named Entity Recognition in Improved transformer with R-Drop structure}





\author[1]{\fnm{Weidong} \sur{Ji}}\email{kingjwd@126.com}

\author[1]{\fnm{Yousheng} \sur{Zhang}}\email{13783686643@163.com}
\author[1]{\fnm{Guohui} \sur{Zhou}}\email{zhouguohui@hrbnu.edu.cn}
\author*[1]{\fnm{Xu} \sur{Wang}}\email{wx971025@163.com}

\affil[1]{\orgdiv{College of Computer Information Engineering}, \orgname{Harbin Normal University}, \orgaddress{ \city{Harbin}, \postcode{150025},  \country{China}}}


\abstract{To enhance the generalization ability of the model and improve the effectiveness of the transformer for named entity recognition tasks, the XLNet-Transformer-R model is proposed in this paper. The XLNet pre-trained model and the Transformer encoder with relative positional encodings are combined to enhance the model's ability to process long text and learn contextual information to improve robustness. To prevent overfitting, the R-Drop structure is used to improve the generalization capability and enhance the accuracy of the model in named entity recognition tasks. The model in this paper performs ablation experiments on the MSRA dataset and comparison experiments with other models on four datasets with excellent performance, demonstrating the strategic effectiveness of the XLNet-Transformer-R model.}

\maketitle

\section{Introduction}\label{1 Introduction}
The main task of Named Entity Recognition (NER) is to recognize entities such as names of people, place names, and proper nouns in a corpus. The NER task is the basis for many NLP downstream tasks, including knowledge graphs \cite{ref1}, intelligent questions and answers systems \cite{ref2}, machine translation \cite{ref3,ref4}, and information retrieval \cite{ref5}. In the early stage, named entity recognition tasks were mainly based on rules and templates \cite{ref6,ref7,ref8}, and gradually shifted to statistical machine learning based methods \cite{ref9,ref10,ref11}. With the wide application of deep learning techniques, named entity recognition based on deep learning models \cite{ref12,ref13} is nowadays gradually becoming a mainstream party. Jason et al \cite{ref12} proposed Bidirectional LSTM-CNNs architecture to automatically detect word and character-level features, eliminating the need for most feature engineering; Gregoric et al. \cite{ref14} used multiple independent Bi-LSTM units at the same input end to enhance the model effect. Ma et al \cite{ref15} fused the input layer for words with all matched lexical information based on the Lattice-LSTM model.The emergence of Transformer has rapidly boosted researchers' research on NER \cite{ref16,ref17}. However, it was found that Transformer is not very effective for NER tasks directly due to the inability to obtain relative positional information between words due to the position encoding method \cite{ref18}. Zheng Honghao et al \cite{ref19} proposed an improved Transformer encoder method for named entity recognition, which improved the absolute positional encodings in the Transformer to relative positional encodings. effectively improving the effect of the NER task. li et al \cite{ref20} used BERT to generate word vectors dynamically and experimentally proved that the replacement with dynamic word vectors has better performance. From the research literature, named entity recognition based on deep learning is more effective, but the model generalization ability is not good and often produces overfitting in the training of domain entity recognition, and the vertical domination research and generalization of the model in named entity recognition still deserve in-depth research.

In this paper, we propose the XLNet-Transformer-R model. The XLNet pre-training model is used as the encoder layers, and the Transform-XL feature extractor is used to enhance the feature extraction and fusion ability in long text. The Transformer with relative positional encodings is used as the positional encoder layers so that the model captures the relative positional information of the input sequence and enhances the model's perception of the input sequence, while the R-Drop structure is used to reduce the occurrence of overfitting and enhance the generalization ability of the model.

\section{Related Work}\label{2 Related Work}

\subsection{XLNET pre-training model}\label{2.1 XLNET pre-training model}
After BERT \cite{ref21} was proposed, researchers applied it to the NER task \cite{ref22,ref23} and made a great breakthrough. An important reason why BERT can obtain natural language understanding ability is the use of the Masked Language Model (MLM), but since Mask tokens in MLM training do not appear in the actual inference results, this can lead to errors in prediction. Google proposed XLNet \cite{ref24} to solve this problem. The XLNet fuses the autoencoding training method of BERT and the autoregressive training method of GPT \cite{ref25} and proposes the Permutation Language Model (PLM), which uses a permutation autoregressive training method to overcome the negative effects of Mask tagging in the autoencoding training approach are overcome. To cooperate with this replacement autoregressive training approach, a two-stream attention is proposed, using one attention mechanism for the above information and one for the below information. In addition, XLNet introduces the Transformer-XL recurrence mechanism to enable the model to better access long-distance contextual information and enhance its ability to handle long texts. The permutation language model, the two-stream attention, and the recurrence mechanism together constitute the three main mechanisms of the XLNet model.

\subsection{Relative Positional Encodings}\label{2.2 Relative Positional Encodings} 
Transformer uses fixed absolute positional encodings, and for each input vector, the word embedding vector is summed by the encoder with the absolute positional encoding, as shown in Eq.(1).
\begin{equation}
	c_{i} =e_{i} + p_{i} \label{eq.(1)}
\end{equation}

However, the absolute positional encodings only undertake the task of differentiation and have no direction. In the NER task, the directional pointing between entities is particularly important, and the attention score between the input vector and the vector is as shown in Eq.(2), and Eq.(3) can be obtained by expanding Eq.(2).

\begin{equation}
	A_{ij} = QK^{T} \label{eq.(2)}
\end{equation}
\begin{equation}
	A_{ij} = e_{i} W^{Q} (W^{K})^{T}  e_{l}^{T} + e_{i} W^{Q} (W^{K})^{T}  p_{l}^{T} + p_{i} W^{Q} (W^{K})^{T}  e_{l}^{T} + p_{i} W^{Q} (W^{K})^{T}  p_{l}^{T}  \label{eq.(3)}
\end{equation}

According to Eq. (3), only the last three items contain the position vector, and the second and third items only contain the absolute position of the  $\i$th and $\l$th elements respectively. According to the research in literature \cite{ref26}, the fourth item cannot reflect the relative positional information, so Transformer encoder cannot reflect the relative position of input information.

In this paper, we refer to the work of Shaw et al. \cite{ref27}, which adds relative positional encodings to Transformers. Assuming that the relative distance of the input sequence does not exceed k, the relative positions of the  matrix and the elements of the   matrix sequence are calculated by Eq.(4) and Eq.(5), respectively, and the relative distances between the elements are calculated by Eq.(6).

\begin{equation}
   a_{il}^{K}\in w_{rel(l-i,k)}^{K} \label{eq.(4)}
\end{equation}

\begin{equation}
  a_{il}^{V}\in w_{rel(l-i,k)}^{V} \label{eq.(5)}
\end{equation}
\begin{equation}
    rel(x,k)=\max (-k,\min (k,x))  \label{eq.(6)}
\end{equation}

For two relative positional encoding $a_{il}^{K},a_{il}^{V}$ , through the training to learn their own matrix, as shown in Eq. (7) and Eq. (8).

\begin{equation}
  {{\mathbf{w}}^{K}}=(\mathbf{w}_{-k}^{K},...,\mathbf{w}_{k}^{K})  \label{eq.(7)}
\end{equation}

\begin{equation}
	{{\mathbf{w}}^{V}}=(\mathbf{w}_{-k}^{V},...,\mathbf{w}_{k}^{V})  \label{eq.(8)}
\end{equation}

The constructed relative position matrix is fused with the multi-headed self-attentive mechanism to form a structure with directional information, as shown in Eq. (9) and Eq. (10).
\begin{equation}
	{{A}_{il}}={{Q}_{i}}{{({{K}_{l}}+a_{il}^{K})}^{T}}  \label{eq.(9)}
\end{equation}
\begin{equation}
	{{A}_{il}}={{Q}_{i}}{{({{K}_{l}}+a_{il}^{K})}^{T}}  \label{eq.(10)}
\end{equation}

Where $A_{il}$ is the attention score, the relative positional encoding information is integrated into the multi-head word attention mechanism in the above way, so that the model can capture the relative positional information of the input sequence and enhance the perception ability of the model to the input sequence.

\section{XLNet-Transformer-R model}\label{3 XLNet-Transformer-R model}

In this paper, the XLNet pre-trained model is used as the encoding layer, the Transformer with relative positional encoding is used as the positional encoding layer, and the R-Drop structure is combined to construct the XLNet-Transformer-R model for the named entity recognition tasks. The training data is input embedded and encoded through XLNet, and the relative position is embedded into the feature encodings output of XLNet through the Transformer encoder with relative positional encodings to ensure that there is relative positional information in the features. The R-Drop structure is used to enhance the generalization ability of the model. XLNet - Transformer - R model structure is shown in Figure 1.

\begin{figure}[h]%
	\centering
	\includegraphics[width=0.8\textwidth]{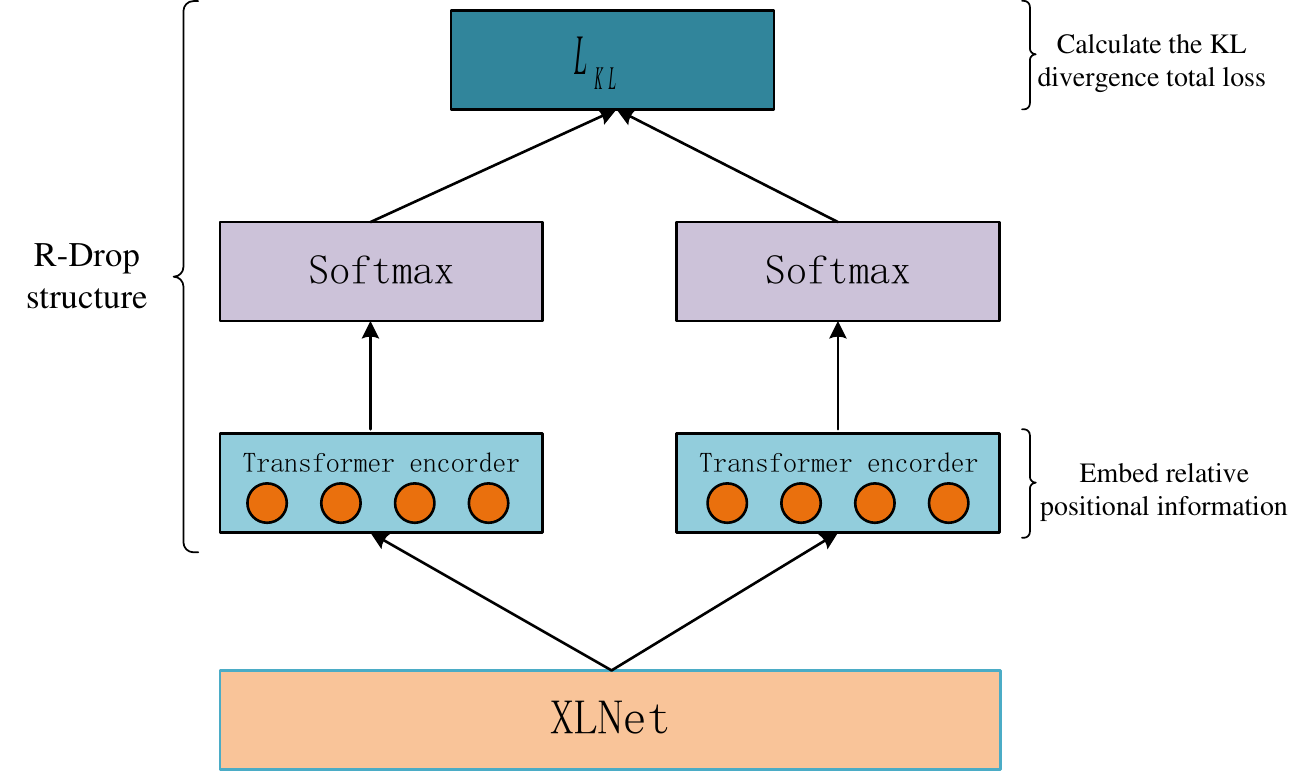}
	\caption{XLNet-Transformer-R model structure \centering}\label{fig1}
		
\end{figure}

By integrating the R-Drop structure into the named entity recognition model, the Transformer model with relative positional encoding has different structures, similar to bagging training, which increases the generalization ability of the model.

\subsection{XLNET Displacement Autoregressive Model}\label{3.1 XLNET Displacement Autoregressive Model}
\subsubsection{Permutation Autoregressionm}\label{3.1.1 Permutation Autoregression}

The mask matrix is used in Transformer to realize the unification of single term coding and bidirectional coding,but the training of unidirectional coding and bidirectional coding are independent of each other in the training process. Permutation Language Model (PLM) skillfully combines one-way encoding training and k-bidirectional encoding training by permutation autoregressive method. This is done by first shuffling the input sequence and then using a one-way encoding mechanism to predict the last 15$\%$ of the input sequence. An example is as follows:

The original order: it's $|$ a $|$ gread $|$ day, 1$|$2$|$3$|$4

The replacement order: gread $|$ a $|$ day $|$ it's, 3$|$2$|$4$|$1

For the word order after permutation, only a special mask matrix is needed to mask, and then one-way coding training is performed. The mask matrix is shown in Figure 2.Although unidirectional encoding can only notice the information of the previous word, the permutation operation can make the current word see its own context information with a high probability, so as to achieve the effect of bidirectional encoding.

\begin{figure}[h]%
	\centering
	\includegraphics[width=0.4\textwidth]{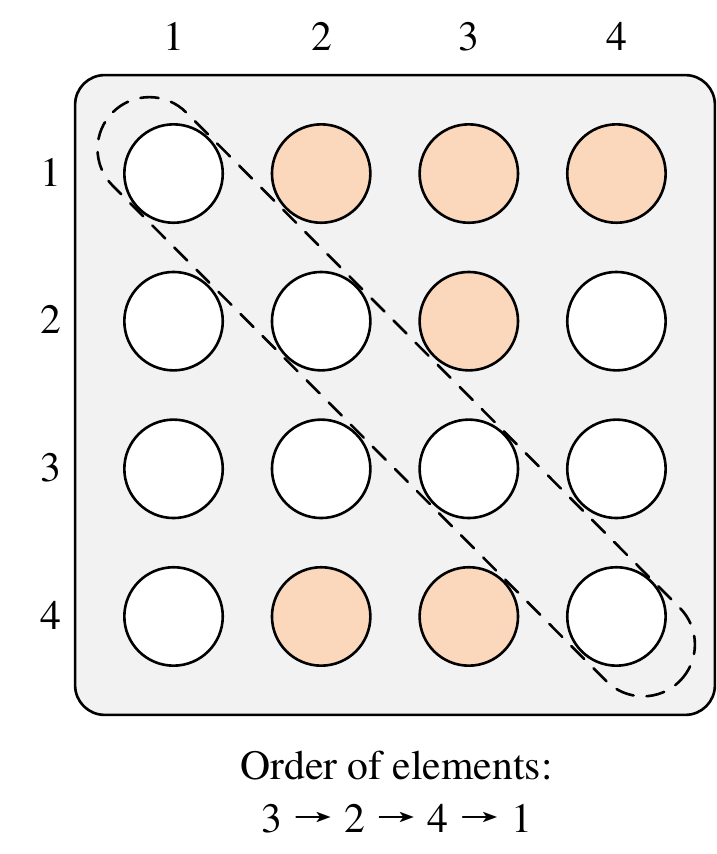}
	\caption{The mask matrix in PLM \centering }\label{fig2}
	
\end{figure}

As can be seen from Figure 2, since position 1 "it’s" is placed on position 4 after permutation, it can see all the previous word information, that is, the orange dot position in the first row of Figure 2. Similarly, since the position of "of" in position 2 remains unchanged after replacement, it can see the content in position 1 after replacement, namely "weather", and so on, the PLM mask matrix can be generated.

\subsubsection{Two-Stream Attention Mechanism}\label{3.1.2 Two-Stream Attention Mechanism}

The mask matrix in PLM is different from the mask matrix in GPT. The mask matrix in PLM masks itself and all the future information. Because the embedded information of the position to be predicted is not encoded, the information disappearance problem arises, for example:

First layer : e3 $|$ e2 $|$ e4 $|$ e1 

Second layer : x3 $|$ x2 $|$ x4 $|$ x1

$\cdots $

N layer : o3 $|$ o2 $|$ o4 $|$ o1

For x3, the encoding information of x3 only contains the information before e3, but does not contain e3 itself. When calculating x2, the information of e3 itself needs to be used, but x3 does not contain the information of e3, so there is a problem of missing information in the encoding of x2. This situation becomes very serious as the number of layers in the network increases. In order to solve this problem, the two-stream attention mechanism is adopted in XLNet to alleviate the above problems.

The two streams refer to content stream and query stream. The content stream is represented by $h$. The mask of content stream is similar to that of Transformer decoder, and the embedded information of the position to be predicted is taken into account when calculating multiple self-attention. The query stream is represented by $g$ , and the mask of the query stream is set according to the PLM training mask. The position to be predicted can only see the previous information, not the information of itself and the future. As shown in Figure 3.

\begin{figure} [h]%
	\centering
	\includegraphics[width=0.8\textwidth]{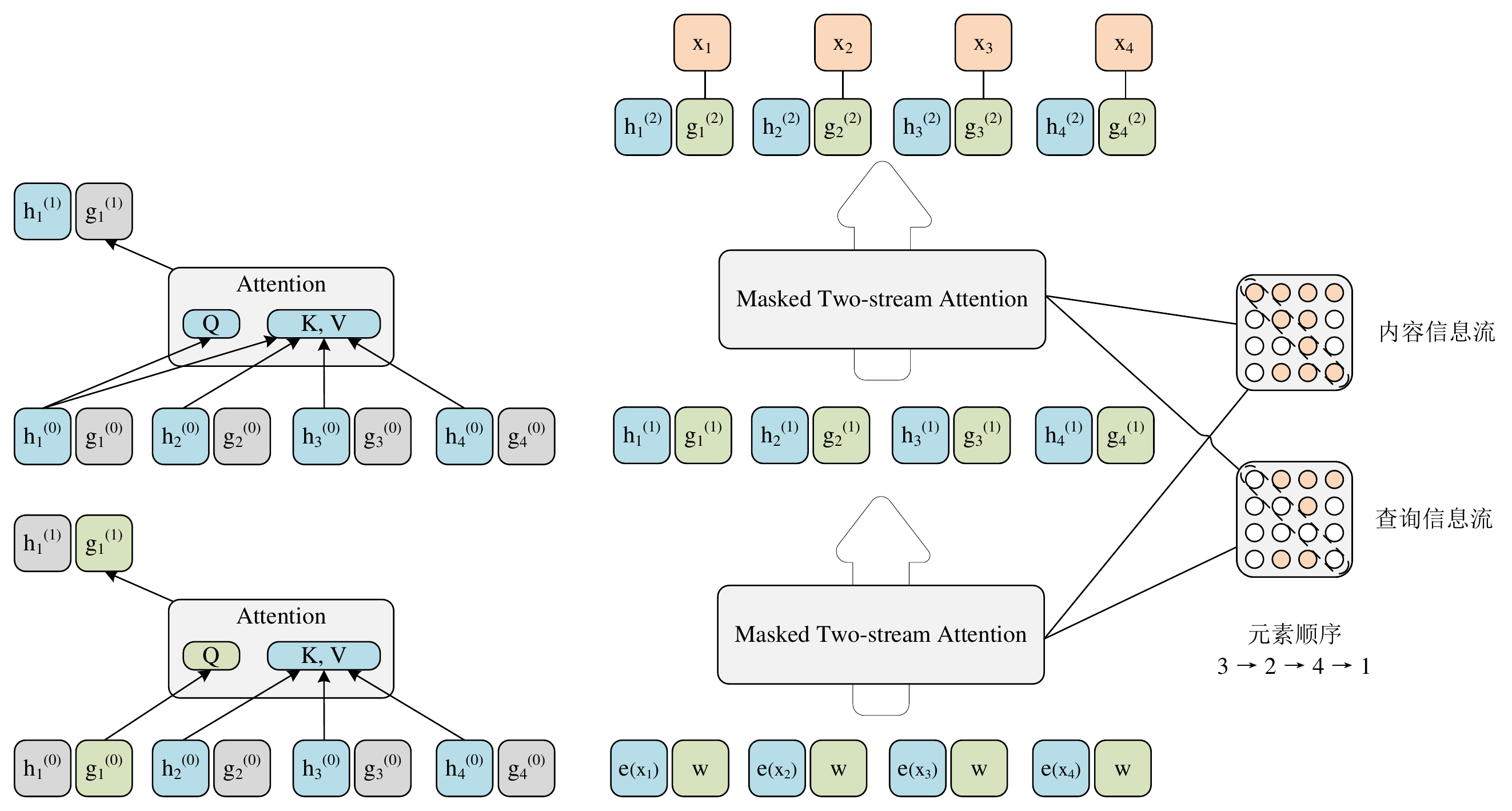}
	\caption{Two-Stream Self-Attention  \centering}\label{fig3}
\end{figure}

It can be seen from the figure that the calculation process of the two-stream self-attention mechanism is: firstly, the initial content stream $h_{i}^{(0)}$ is the word vector embedded in $e({{x}_{i}})$; the initial qurty stream $g_{i}^{(0)}$ is the variable $w$; the superscript represents the 0th layer encoder; the encoder calculation is shown in equations (11) and (12).

\begin{equation}
		g_{{{z}_{t}}}^{(m)}=Attention(Q=g_{{{z}_{t}}}^{(m-1)},KV=h_{{{z}_{<t}}}^{(m-1)}; \label{eq.(11)}
\end{equation}

\begin{equation}
	h_{{{z}_{t}}}^{(m)}=Attention(Q=h_{{{z}_{t}}}^{(m-1)},KV=h_{{{z}_{<=t}}}^{(m-1)}; \label{eq.(12)}
\end{equation}

Where ${{z}_{t}}$ represents the current position, ${{z}_{<t}}$ represents the forward information without the current position, and ${{z}_{<=t}} $ represents the forward information including the current position. When calculating the content information flow, ${{h}_{i}}$ uses the same multi-head self-attention calculation method as the Transformer decoder, and the encoding carries the character embedding information of the current position. ${{g}_{i}}$ does not carry its own embedding information when calculating the encoding, but only carries its own position information. The dual-stream self-attention mechanism saves content information through an additional information preservation stream, solving the problem of information loss in PLM training.

\subsubsection{Transformer-XL}\label{3.1.3 Transformer-XL}

Transformer-XL (Transformer eXtra Long) is a model proposed by Google in 2019 to address the issue of processing long texts in BERT. Compared to RNN, Transformer can capture long-term dependencies, but it requires more intermediate information to be stored, which increases with sequence length. If the text is too long, Transformer can quickly cause memory overflow. Due to this performance issue, BERT splits continuous text into segments of 512 words, which solves the memory issue but severely damages the long-term dependency relationships within the text. For example, The little girl is holding ice cream |her mother bought it for her ,if "the little girl" and "her" are not in the same segment, the encoding of "her" will not be able to merge with the information of "the little girl". Transformer-XL introduces a recurrent state mechanism similar to RNN, first dividing the text into segments of fixed length 512, introducing a state variable "memory", and saving the encoding information of the previous segment into memory, which is used for the encoding of the next segment. This way, the information from the previous segment can be captured by the next segment, allowing the paragraph information to be transmitted in the encoder across time steps and obtaining long-term dependency relationships. The transmission of information between segments is shown in Figure 4.

\begin{figure} [h]%
	\centering
	\includegraphics[width=0.8\textwidth \centering]{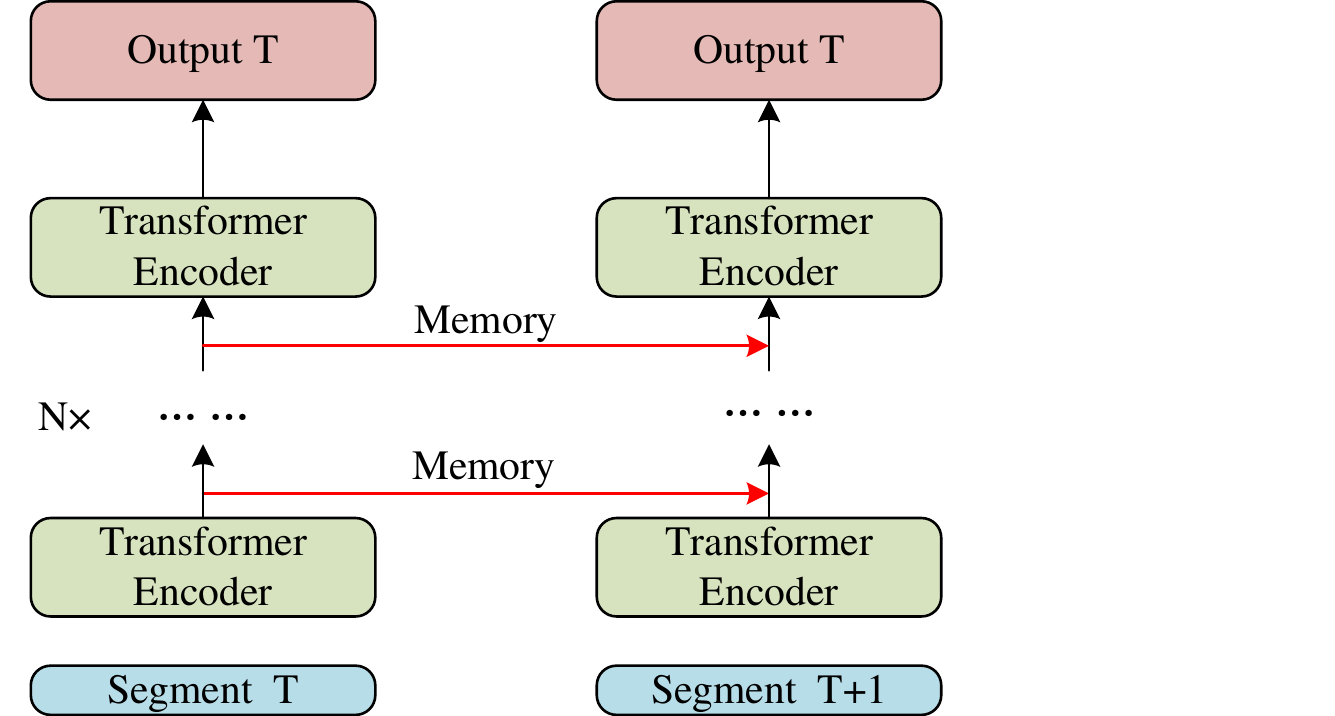}
	\caption{Transformer-XL inter-segment information transfer \centering}\label{fig4}
\end{figure}

After introducing the segment recurrence mechanism, Transformer-XL encountered the issue of position encoding. In Transformer, absolute position encoding is used, where the position encoding for each segment is the same for the same position. However, in the segment recurrence mechanism, the position of the first word in paragraph 1 and the position 1 word in paragraph 2 have a temporal relationship, and the position encoding for the same position in two different segments cannot be the same. Therefore, Transformer-XL uses sine relative positional encoding.

XLNet uses the segment recurrence mechanism from Transformer-XL and can capture long-term dependencies compared to the BERT model. The illustration of how XLNet captures long-term dependencies is shown in Figure 5.

\begin{figure} [h]%
	\centering
	\includegraphics[width=0.8\textwidth]{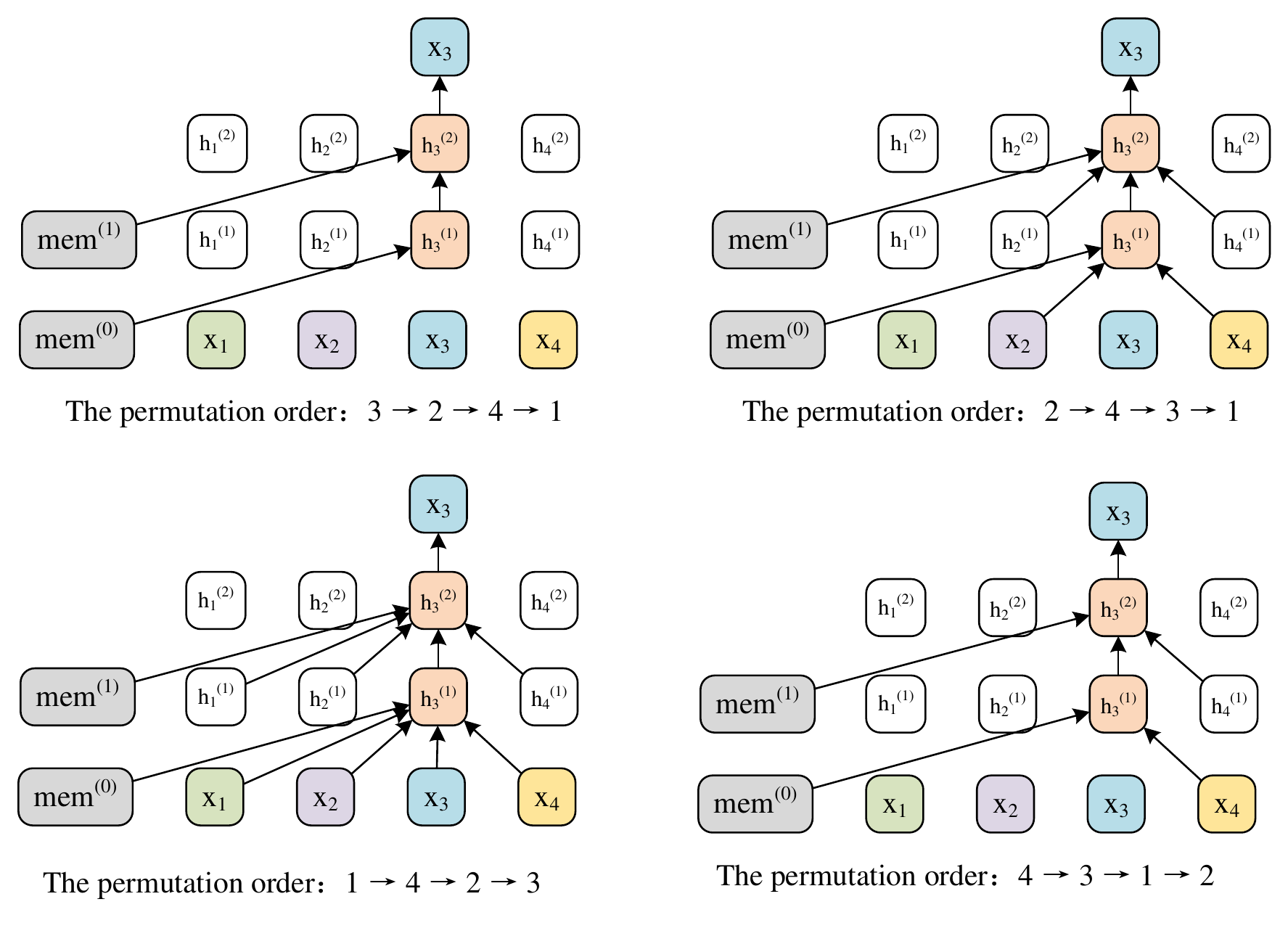}
	\caption{XLNet captures long-term dependencies \centering}\label{fig5}
\end{figure}

The XLNet model greatly improves the feature extraction and fusion capabilities in long texts by integrating the feature extractor from Transformer-XL.

\subsection{R-Drop structure}\label{3.2 R-Drop structure}

Named entity recognition based on deep learning is excellent, but the generalization ability of the model is not good, which often produces overfitting in the training of domain entity recognition. To improve this situation, the R-Drop structure \cite{ref28} will be introduced in this paper to enhance the generalization ability of the model, and the R-Drop structure is shown in Figure 6.

\begin{figure}[ht]%
	\centering
	\includegraphics[width=0.8\textwidth]{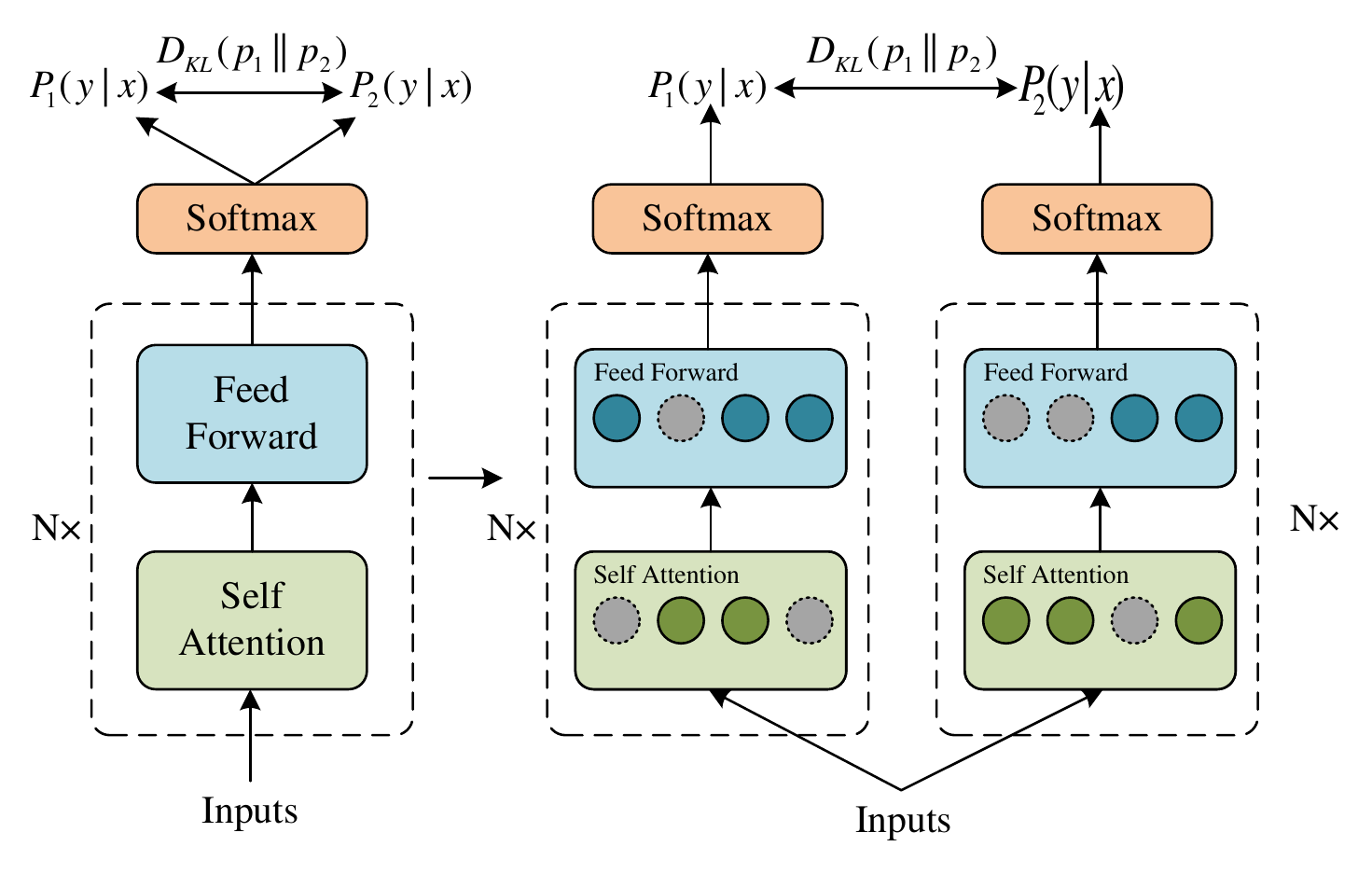}
	\caption{R-Drop structure  \centering}\label{fig6}
\end{figure}

For the same training data, feeding the data into two different Dropout structures results in different data distributions, approximating the two paths as two different models. Multiple outputs obtained through R-Drop are first used to compute cross-entropy loss as shown in Equation (13), and then total loss can be calculated using KL divergence, which is shown in Equation (14).

\begin{equation}
	L_{i}^{CE}=-\log P_{1}^{\theta }({{y}_{i}}|{{x}_{i}})-\log P_{2}^{\theta }({{y}_{i}}|{{x}_{i}}) \label{eq.(13)}
\end{equation}

\begin{equation}
	(P_{1}^{\theta }({{y}_{i}}|{{x}_{i}})||P_{2}^{\theta }({{y}_{i}}|{{x}_{i}}))+KL(P_{2}^{\theta }({{y}_{i}}|{{x}_{i}})||P_{1}^{\theta }({{y}_{i}}|{{x}_{i}})) \label{eq.(14)}
\end{equation}

The KL divergence serves to hope that the outputs of different Dropout models are as consistent as possible. Finally, the final Loss can be obtained by weighting the two Losses, as shown in equation (15).

\begin{equation}
	{{L}_{i}}=L_{i}^{CE}+\alpha L_{i}^{KL} \label{eq.(15)}
\end{equation}

In the actual training process, the method used is not to feed the same input into the same model twice. To save training time, the common practice is to copy the training data directly, and the batch$\_$size will be directly doubled, which can greatly save training time, but the model complexity rises rapidly, and the training time of each step will increase, but the total time will be shortened.

\section{Experimental design and result analysis}\label{4  Experimental design and result analysis}
\subsection{Dataset and model parameter settings}\label{4.1 Dataset and model parameter settings}

This experiment selects MSRA, Resume, OntoNotes4.0 and Weibo datasets for general domain NER task training, and uses the entity set in the data structure direction open-sourced by the School of Data Science, East China Normal University for vertical domain NER task training. The MSRA \cite{ref29} dataset is a public dataset from Microsoft, including three types of entities: personal name, place name, and organization name. Resume \cite{ref30} is a Chinese resume dataset, containing eight types of entities including country, education institution, personal name, organization name, place name, occupation, position, and ethnicity. OntoNotes4.0 \cite{ref31} is a large-scale multilingual NER training dataset, and this article only takes the Chinese part of the dataset. The Weibo dataset \cite{ref32} is a social media dataset that contains four types of entities: place name, personal name, organization name, and geopolitical. The NER training set open-sourced by the School of Data Science, East China Normal University, includes named entities and relations in middle school mathematics, high school mathematics, and data structures. This article selects the data structure direction training set for training. The detailed information of the training set is shown in Table 1.

\begin{table}[h]
	\caption{Detailed information of named entity recognition training set  \centering}\label{tb1}%
	\begin{tabular}{|p{2cm}|p{2cm}|p{2cm}|p{2cm}|p{2cm}|}
		\hline
		Dataset & The total sentence & Number of sentences in the training set & Number of sentences in the validation set & Number of sentences in the test set  \\
		\hline
		MSRA          & 46400  & 46400 & -    &  -   \\
		\hline
		Resume        & 4740   & 3800  & 460  & 480  \\
		\hline
		OntoNotes4.0  & 24300  & 15700 & 4300 & 4300 \\
		\hline
		Weibo         & 1940   & 1400  & 270  & 270 \\
		\hline
		Open source dataset of East China Normal University &176919  & 66497 & -  & 110422 \\
		\hline
	\end{tabular}
\end{table}

Data set labeling rules all use BMES labeling rules, where B is the beginning position of the entity, M is the middle of the entity, E is the end of the entity, and S is the single-word entity.

The experimental model uses the XLNet-Transformer-R structural model, and the XLNet pre-trained model using the XLNet model pre-trained on encyclopedia and news data using 5.4B word count in Xunfei Lab of Harbin Institute of Technology is used as the embedding layer. A 12-layer self-attentive mechanism with 8 heads is used in the Transformer with relative positional encoding, and the maximum relative encoding distance of relative position encoding is dynamically adjusted from 1 to 15. The model is optimized using the Adam optimizer with a preheated learning rate mechanism. The learning rate is set to an initial score of 0.002, and the Dropout of the two R-Drop branches is set to 0.85. The training data are collated and the experimental results are compared on all comparison models.

\subsection{Analysis of experimental results for generic domain data training}\label{4.2 Analysis of experimental results for generic domain data training}

In this paper, ablation experiments are first conducted on the XLNet-Transformer-R model on the MSRA dataset, and the ablation experimental results are obtained by comparing the model without R-Drop structure, the model without relative position encoding and other encoder models, and the results show that the encoder, relative position encoding strategy of the XLNet-Transformer-R model and R-Drop structure are effective. The experimental results are shown in Table 2.

\begin{table}[h]
	\caption{XLNet-Transformer-R ablation experiments  \centering}\label{tb2}
	\begin{tabular}{|c|c|c|c|}
		\hline
		Model & Precise $(\%)$  & Recall $(\%)$ & F1 Score $(\%)$  \\
		\hline
		XLNet-GRU     & 88.23  & 88.19 & 88.21   \\
		\hline
		XLNet-BiLSTM  & 89.98  & 86.75  & 88.34   \\
		\hline
		XLNet-Transformer(Absolute positional encodings)
		  & 88.56  & 86.11 & 87.32  \\
		\hline
		XLNet-Transformer  (Relative positional encodings)
		  & 90.23   & 93.57  & 91.87  \\
		\hline
		XLNet-Transformer-R  (Absolute positional encodings)
		 & 90.02  & 87.39   & 88.69 \\
		\hline
		XLNet-Transformer-R	(Relative positional encodings) (this paper) 
		& \textbf{92.07} & \textbf{94.01}     &\textbf{93.03}	\\
		\hline
	\end{tabular}
\end{table}

From the ablation experimental results, compared with the XLNet-Transformer model without relative positional encodings and R-Drop structure, the F1 score of the model in this paper is improved by 5$\%$, and the addition of relative positional encodings strategy improves the F1 score by about 4$\%$, and the addition of R-Drop structure improves the F1 score by about 2$\%$, which proves the XLNet-Transformer-R model's strategy effectiveness.

This paper compared the XLNet-Transformer-R model with the Lattice LSTM model [30], the FLAT model \cite{ref33}, the XLNet-Transformer$\_$P-CRF model \cite{ref19}, and the model described in Reference \cite{ref34} on four datasets. The comparative experimental results on the four datasets are shown in Table 3, Table 4, Table 5, and Table 6.

\begin{table}[h]
	\caption{Comparative experiments on the MSRA dataset  \centering }\label{tb3}
	\begin{tabular}{|c|c|c|c|}
		\hline
		Model & Precise $(\%)$  & Recall $(\%)$ & F1 Score $(\%)$  \\
		\hline
		Lattice LSTM    & 93.57  & 92.79 & 93.18   \\
		\hline
		Reference [34]  & 94.01  & 92.93  & 93.47   \\
		\hline
		FLAT            & -  & - & 94.35  \\
		\hline
		XLNet-Transformer$\_$P-CRF  & \textbf{93.64}  & \textbf{96.63} & \textbf{95.11}  \\
		\hline
		XLNet-Transformer-R & 92.07  & 94.01   & 93.03 \\
		\hline
	\end{tabular}
\end{table}

\begin{table}[h]
	\caption{Comparison experiments on Resume dataset  \centering }\label{tb4}
	\begin{tabular}{|c|c|c|c|}
		\hline
		Model & Precise $(\%)$      & Recall $(\%)$ & F1 Score $(\%)$  \\
		\hline
		Lattice LSTM                & 94.81  & 94.11 & 94.46   \\
		\hline
		FLAT                        & -  & - & 94.93  \\
		\hline
		XLNet-Transformer$\_$P-CRF  & 96.23   & 97.18 & 96.70  \\
		\hline
		XLNet-Transformer-R         &\textbf{96.45}&\textbf{97.23}  &\textbf{96.84} \\
		\hline
	\end{tabular}
\end{table}

\begin{table}[h]
	\caption{Comparative experiments on OntoNotes4.0 dataset  \centering }\label{tb5}
	\begin{tabular}{|c|c|c|c|}
		\hline
		Model & Precise $(\%)$      & Recall $(\%)$ & F1 Score $(\%)$  \\
		\hline
		Lattice LSTM                & 76.35  & 71.65 & 73.88   \\
		\hline
		Reference [34]             & 75.06    &74.52 &74.79 \\
		\hline
		FLAT                        & -       &   - & 75.70  \\
		\hline
		XLNet-Transformer$\_$P-CRF & 76.68   & \textbf{84.80} & 80.54  \\
		\hline
		XLNet-Transformer-R        & \textbf{79.34} & 83.16  &\textbf{81.21} \\
		\hline
	\end{tabular}
\end{table}

\begin{table}[h]
	\caption{Comparative experiments on microblog datasets  \centering }\label{tb6}
	\begin{tabular}{|c|c|c|c|}
		\hline
		Model & Precise $(\%)$      & Recall $(\%)$ & F1 Score $(\%)$  \\
		\hline
		Reference [34]             & 67.31    &48.61 &56.45\\
		\hline
		Lattice LSTM                & -       & -   & 58.79   \\
		\hline
		FLAT                        & -       & -     & 63.42  \\
		\hline
		XLNet-Transformer$\_$P-CRF  & 68.34  & \textbf{74.88} & 71.46  \\
		\hline
		XLNet-Transformer-R       & \textbf{70.23} & 74.68  & \textbf{72.39} \\
		\hline
	\end{tabular}
\end{table}

From the above experimental results, the XLNet-Transformer-R model is superior to the Lattice LSTM model, the FLAT model, and the model in reference [34]. It has higher precision and F1 score than the XLNet-Transformer$\_$P-CRF model. The F1 Score bar chart of the algorithms is shown in Figure 7.

\begin{figure}[ht]%
	\centering
	\includegraphics[width=0.8\textwidth]{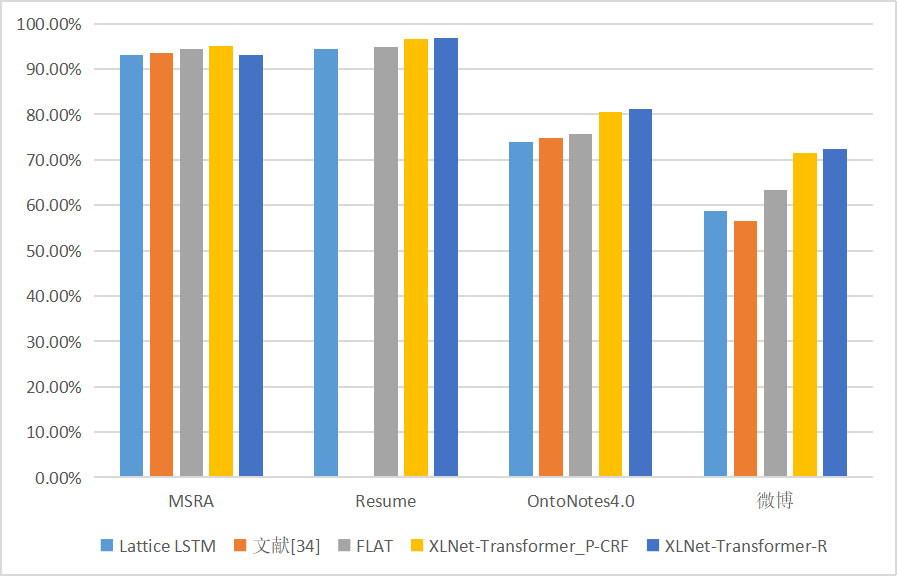}
	\caption{Comparison experimental bar chart  \centering}\label{fig7}
\end{figure}

From the figure, we can see that the XLNet-Transformer-R model performs well on the four test sets, which proves the effectiveness of the XLNet-Transformer-R model.
\subsection{Analysis of experimental results of vertical domain data}\label{4.3 XLNET Displacement Autoregressive Model}

In this paper, the XLNet-Transformer-R model is trained using the publicly available data structure direction NER training set from the Data Institute of East China Normal University, and the data in the dataset are labeled in BIO, and this paper first preprocesses the dataset to convert the data to BMES labeling. The same parameter settings as those used for training the generic domain data are used, and the results of the comparison experiments are shown in Table 7.

\begin{table}[h]
	\caption{Comparison of experimental results of vertical domain data training  \centering }\label{tb7}
	\begin{tabular}{|c|c|c|c|}
		\hline
		Model & Precise $(\%)$      & Recall $(\%)$ & F1 Score $(\%)$  \\
		\hline
		Lattice LSTM                & 91.30      &90.28  & 90.79   \\
		\hline
		BiLSTM-CRF  & 90.76   & 88.47 & 89.23 \\
		\hline
	    XLNet-Transformer-R(this paper)   & \textbf {95.83} &\textbf{92.89} & \textbf{94.34} \\
		\hline
	\end{tabular}
\end{table}

As can be seen from Table 7, the F1 score of the XLNet-Transformer-R model in the data structure direction test set are better than the other two comparison models, which proves that the XLNet-Transformer-R model still has good performance in the vertical domain.

\section{Conclusion}\label{5 Conclusion}
This article proposes the XLNet-Transformer-R model. The XLNet pre-training model is used as the encoding layer to solve the mutual dependency between Mark words. The inclusion of Transform-XL allows for the retrieval of longer distance information. The relative positional encodings is integrated into the multi-head self-attention mechanism to capture the relative positional relationships of the input sequence and enhance the model's perception ability. The R-Drop structure is introduced to enhance the model's generalization ability. Through ablation and comparative experiments on a general domain dataset, the effectiveness of the XLNet-Transformer-R model strategy is demonstrated. Comparative experiments on a vertical domain dataset also demonstrate that the model performs well in vertical domains.

\section*{Declarations}
\bmhead*{Authors:}  
\   

1. Weidong Ji, Male, born in 1978, Doctor, Professor, research direction is swarm intelligence, big data.

2. Xu Wang, Male, born in 1997, master’s degree, research direction is swarm intelligence.

3. Yousheng Zhang, Male, born in 1998, master’s degree, research direction is named entity recognition.

4. Guohui Zhou, Male, born in 1973, Doctor, Professor, research direction is swarm intelligence artificial intelligence.

\bmhead*{Data availability:}
\

Te datasets used in this research work are publicly available and can be downloaded from the websites below.
\

MSRA:\url{https://github.com/Houlong66/lattice_lstm_with_pytorch/tree/master/data}
\

Resume: \url{https://github.com/luopeixiang/named_entity_recognition/tree/master/ResumeNER}
\

OntoNotes4.0:\url{https://paperswithcode.com/accounts/login?next=/dataset/ontonotes-4-0}
\

Weibo:\url{https://github.com/Houlong66/lattice_lstm_with_pytorch/tree/master/data} 

\bmhead*{Editorial Board Members and Editors:}
\

The author has no competitive interest with members of the Editorial Committee and Editors.

\bmhead*{Funding:}
\

1. the National Natural Science Foundation of China(31971015)

2. Natural Science Foundation of Heilongjiang Province in 2021(LH2021F037)

\bmhead*{Competing financial interests:}
\

The authors have no competing interests to declare that are relevant to the
content of this article.

\bibliography{sn-bibliography}

\end{document}